\documentclass{article}
\usepackage{spconf,amsmath,epsfig}

\usepackage{spconf,amsmath,epsfig}
\usepackage{graphicx}
\usepackage{booktabs}
\usepackage{float}

\usepackage{graphicx}
\usepackage{float}
\usepackage{subfig}
\usepackage{cite}
\usepackage{url}
\usepackage{amssymb}
\usepackage{bm}
\usepackage{epstopdf}

\usepackage{graphicx}
\usepackage{float}
\usepackage{subfig}
\usepackage{cite}
\usepackage{url}
\usepackage{amssymb}
\usepackage{bm}
\usepackage{epstopdf}

\begin{document}\sloppy

\def\x{{\mathbf x}}
\def\L{{\cal L}}

\title{Non-rigid 3D shape retrieval based on multi-view metric learning}
%
\name{ Haohao Li, Shengfa Wang, Nannan Li, Zhixun Su, Ximin Liu }
\address{ School of Mathematical Sciences, Dalian University of Technology, Dalian, Liaoning, China, 116024 }

\maketitle

\begin{abstract}
This study presents a novel multi-view metric learning algorithm, which aims to improve 3D non-rigid shape retrieval. 
With the development of non-rigid 3D shape analysis, there exist many shape descriptors. The intrinsic descriptors can be explored to construct various intrinsic representations for non-rigid 3D shape retrieval task. The different intrinsic representations (features) focus on different geometric properties to describe the same 3D shape, which makes the representations are related. Therefore, it is possible and necessary to learn multiple metrics for different representations jointly. We propose an effective multi-view metric learning algorithm by extending the Marginal Fisher Analysis (MFA) into the multi-view domain, and exploring Hilbert-Schmidt Independence Criteria (HSCI) as a diversity term to jointly learning the new metrics. The different classes can be separated by MFA in our method. Meanwhile, HSCI is exploited to make the multiple representations to be consensus. The learned metrics can reduce the redundancy between the multiple representations, and improve the accuracy of the retrieval results. Experiments are performed on SHREC'10 benchmarks, and the results show that the proposed method outperforms the state-of-the-art non-rigid 3D shape retrieval methods.

\end{abstract}
\section{Introduction}
\label{sec:intro}
With the rapid development of 3D model acquirement technique, massive amounts of 3D shape data had been captured and many large-scale 3D shape datasets were created, such as Google 3D warehouse, Shapenet, etc. The 3D shape analysis has been an active research spot for decades. Among numerous challenges in 3D shape analysis, non-rigid 3D shape retrieval \cite{wu2019cycle} is one of the most important problems. The 3D shape retrieval can be simply described as follows: Given a query shape, we would like to develop a algorithm to determinate the similarity of the query shape to all the shapes in a large collection of 3D shapes. Most conventional non-rigid 3D shape retrieval approaches compute the similarity between the shapes based on a hand-crafted representation and metric. Once the representations of the shapes are obtained, one can explore a metric to measure the similarity of the representations for retrieval task. However, only one type representation usually can not contain enough geometric structure information to characterize the shapes with non-rigid deformations well. Therefore, making full use of the geometric properties by multiple representations \cite{wu20193D} to improve the shape retrieval \cite{wang2017effective} is a hot topic.

In the past years, various approaches for non-rigid 3D shape retrieval has been proposed \cite{BronsteinBGO11,LitmanBBC14,xie2015deepshape,Ghodrati2017Nonrigid}. Most of these approaches mainly focus on construction of the intrinsic representation  and calculation of similarity between the representations \cite{wu2019deepattention}.
The intrinsic representations are usually constructed based on the intrinsic descriptors by using various approaches such as BoW (Bag of Words) and locality constrained linear coding (LLC).
The past decade a large number of shape intrinsic descriptors have been extracted based on Laplace-Beltrami operator for non-rigid 3D shape analysis, such as ShapeDNA \cite{ReuterWP06}, Heat Kernel Signature ($HKS$) \cite{SunOG09}, scale-invariant Heat Kernel Signature ($siHKS$) \cite{BronsteinK10} and Wave Kernel Signature ($WKS$) \cite{AubrySC11}, etc. A comprehensive survey in \cite{LiH14} has provided details of these spectral signatures.  In \cite{BronsteinBGO11}, the authors utilized bag-of-words (BoW) to construct the representation. A set of spectral signatures such as $HKS$, $WKS$, etc., were clustered by k-means to construct the dictionary of words. Then they added the spatial information into the histogram as shape representation. At last, Similarity Sensitive Hashing (SSH) was explored to improve the performance of the retrieval. Unlike the standard BoW methods, the authors of \cite{LitmanBBC14} proposed a supervised learning of BoW approach with weighted mean pooling to form the representation. In \cite{xie2015deepshape}, it explored auto-encoder to optimize the histogram of $HKS$. These methods pay more attention to the construction of the representation, and the similarity between the rpresentation usually based on Euclidean metric. Some methods utilized metric learning method to get a new metric for improving their performance.  
The authors of \cite{Chiotellis2016Non} obtained the representation by utilizing a weighted mean pooling on point signatures, and linearly mapped into subspace by Large Margin Nearest Neighbor (LMNN) algorithm directly. These methods usually utilize one intrinsic descriptor to form the representation, which can not contain enough geometric information for 3D shape retrieval on large-scale 3D shape datasets. Multiple descriptors should be utilized to tackle this problem. Although descriptors from different views reflect different properties of one same shape, they contain compatible and complementary geometric information which can improve the retrieval performance. And the methods that explored multiple descriptors to construct the hiotellis2016Nonshape representation. For example, in \cite{OhbuchiF10}, the authors compute the weighted average of the distance between the representations constructed by single descriptor. However, these concatenating multiple features methods is not physically meaningful, and caused over-fitting on a small training sample \cite{Chang2013A}.  Thus we develop an multi-view metric learning method combining multiple types representation to achieve the improvement of performance compared to single-view methods for non-rigid 3D shape retrieval.

Generally speaking, metric learning methods \cite{wu2018andwherewhen} usually aim to learn a metric function so that the distance between the data under new metric can be consistent with their labels. In past years, a large number of metric learning methods have been proposed \cite{wang2015robust,feng2018learning,wang2016semantic,Wang2016Iterative,wang2018multiview} such as PCA , LMNN , SVMs , ITML , LDA , LPP etc., which are widely utilized in real-word applications. Meanwhile, a great many methods of learning from multi-view data have also been proposed \cite{Christoudias2012Multi}. Single-view methods always drop some information of the data, which affects the performance. Multi-view learning methods \cite{wang2016multi} can tackle this problem by considering different views jointly, which can fully exploit information from multiple views to improve the performance.  Multi-view learning methods mainly exploit consensus principle or complementary principle to improve learning performance.  You can read a survey about multi-view learning methods in \cite{Chang2013A}.
In this paper, we propose a novel multi-view metric learning method for non-rigid 3D shape retrieval, which can fully utilize the geometric information by jointly learning multiple metrics. To this end, we develop a multi-view metric algorithm named Marginal Fisher Analysis Multi-view Metric learning (MFAMML). We utilize MFA to seek for the optimal projections, which considers both local manifold structure and label information. Since projections can maintain the local manifold structure and maximize the class separability simultaneously, it  can potentially achieve higher performance than Euclidean distance metrics in original feature space. Furthermore, the Hilbert-Schmidt Independence Criterion (HSIC) is employed measure the dependence among different views, which can co-regularize different views and enforce them to be consistent in kernel space. With HSIC term, we can explore the compatible and complementarity information from different views. At last, the alternating maximization is carried out for optimizing the proposed approach. Figure1. shows the pipeline of the proposed framework.

\begin{figure*}[!htb]
	\centering
	\includegraphics[width=1\linewidth]{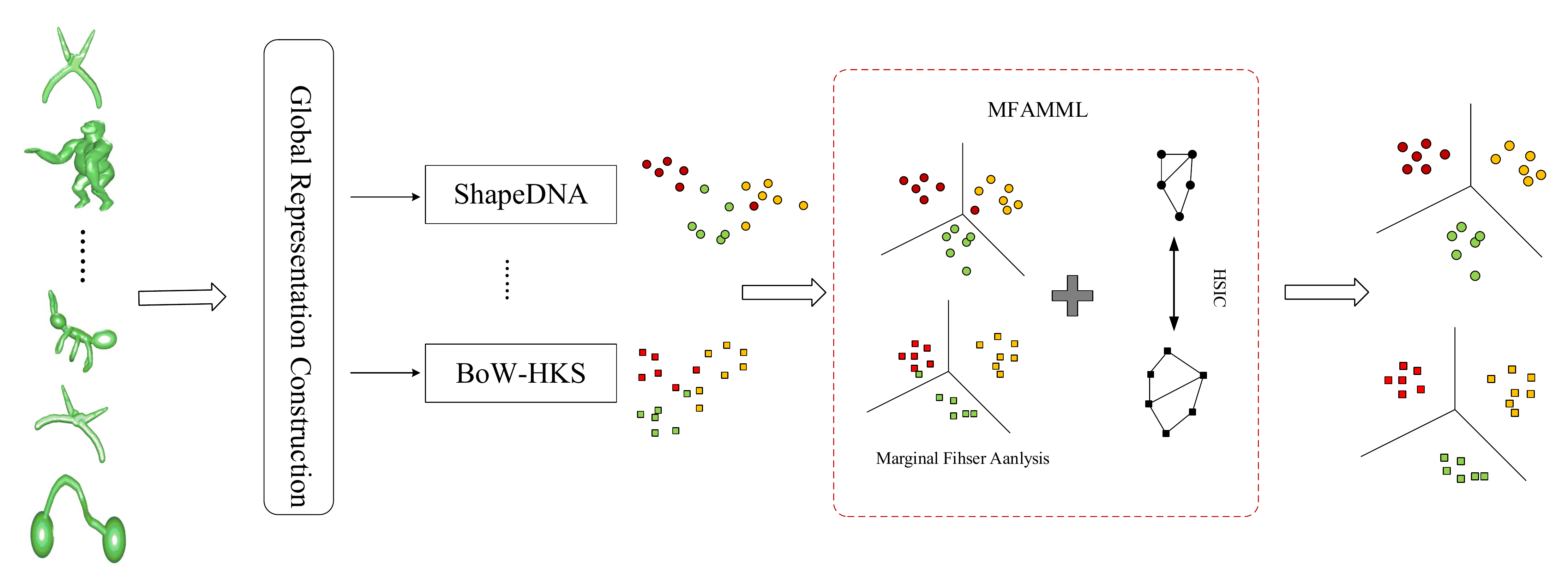}
	\caption{pipeline of Our proposed framework.}
	\label{fig:1}	
\end{figure*}

\section{Background}
In this section, we briefly review the concept of Mahalanaobis distance. Similar to most of the metric learning methods, which aims to seek Mahalanobis distance for specific application, we also focus on find the Mahalanobis distance functions.
Generally, given a sample set $X = [x_{1},x_{2},...,x_{1=N}] \in R^{d \times N}$ with $N$ samples and $x_{k} \in R^{d\times 1}$, the definition of the Mahalanobis distance between $x_{i}$ and $x_{j}$ is as:
$$
d_{M}(x_{i},x_{j}) = \sqrt{(x_{i}-x{j})^{T}M(x_{i}-x_{j})}
$$
The $d_{M}(x_{i},x_{j})$ should be satisfied the properties of nonnegativity, symmetry, identity of indiscernible, and triangle inequality, which are from definition of distance function. Hence, $M$ must be a symmetric PSD matrix, and can be decomposed by as:
$$
M = WW^{T}
$$
where $W\in R^{ d\times p }$. Therefore, learning a Mahalanobis distance metric $M$ is equivalent to seeking a linear projection $W$, which projects the sample $x_{i}$ to a new space. The learned Mahalanobis metric is equal to the Euclidean metric in the new space.

\section{Proposed Approach}

In this section, we present the detail of the MFAMML method for non-rigid 3D shape retrieval. We extract different types of 3D shape descriptors to construct representations. Some global descriptors that contain the global geometric structure information can be used as the representation directly, such as ShapeDNA, Modal Function Transformation (MFT), etc. For some point signatures such as $HKS$, $siHKS$,  $WKS$, etc. should be encoded by BoW method to form the representations of the shapes. Once the multiple representations are constructed, we use these representations as inputs to learning the metrics for every representation jointly. For single view, the new metric maximize the between-class separability. Meanwhile, every representation is enforced to be consistent by HSIC term, which makes the multiple representations to be more discriminative under the new metric by integrating the compatible and complementarity information from different representation. At last, every representation is projected in a new space and concatenated for shape retrieval.

\subsection{The Process of MFAMML}
We first introduce the MFA algorithm \cite{Dong2007Marginal} for a single view representation $X^{v}$.
Let $X^{v} = [x^{v}_{1},x^{v}_{2},...,x^{v}_{N}],\  x^{v}_{i}\in R^{n_v},\  v = 1,2,...,m$, denote the $v-$th view of $N$ samples, and the dimension of it is $n_v$.
Let $G_{v} = \{X_{v},S_{v}\}$ be the similarity graph with the vertex set $X^{v}$ and similarity matrix $S_{v}$. And the corresponding degree matrix $D_{v}$ and laplace matrix $L_{v}$ are as follows:
$$L_{v}=D_{v}-S_{v}, D^{ii}_{v} = \sum_{j\neq i} S^{ij}_{v}\quad   \forall i.$$

The graph embedding algorithm is to compute a low-dimensional representation of the vertex  set, which can preserves the similarities between pairs of data point in orginal space. Standard graph embedding algorithm aims to find the low-dimensional embedding of vertices as $Y^{v} = [y^{v}_{1},y^{v}_{2},...,y^{v}_{N}]$, which can maintain the similarities among the vertex pairs:
\begin{equation}
\begin{split}
 Y^{v}_{*} = \underset {Y^{v}}{argmin} Tr(Y^{v}L_{v}(Y^{v})^{T}) \\
\text {with} \quad Tr(Y^{v}B_{v}(Y^{v})^{T})=c.
\end{split}
\end{equation}
where $c_{v}$ is a constant and $Tr(A)$ is the trace of matrix $A$. Matrix $B_{v}$ is the constraint matrix based on some more general constraints among the vertices.

The MFA fit the linearization formulation of the
graph embedding framework. It considers the label information into the graph embedding, and aims to obtain a projection with considering both local manifold structure and label information.  
The objective function of it is defined as:
\begin{equation}
\begin{split}
W^{v}_{*} = \underset {W^{v}}{argmax} Tr((W^{v})^{T}X^{v}B_{v}(X^{v})^{T}W^{v}.\\
\text{with}\quad Tr((W^{v})^{T}X^{v}L_{v}(X^{v})^{T}W^{v})=c_{v}
\end{split}
\end{equation}
where the $W^{v}\in R^{n_v} \times d$ is the projection matrix.
In this version, the $B$ and $L$ are defined as:

$$ S_{v}^{ij} =
\begin{cases}
1,&  \text{if}\ i\in N_{k_{1}}(j)\  \text{or}\ j \in N_{k_{1}}(i)\\
0,& \text{else}
\end{cases}$$

$$ \overline{S_{v}^{ij}} =
\begin{cases}
1,&  \text{if}\ (i,j)\in P_{k_{2}}(l(x^{v}_{j}))\  \text{or}\ (i,j) \in P_{k_{2}}(l(x^{v}_{i}))\\
0,& \text{else}
\end{cases}$$

$$L_{v}=D_{v}-S_{v}\ D^{ii}_{v} = \sum_{j\neq i} S^{ij}_{v}\quad \forall i $$

$$B_{v}=\overline{D_{v}}-\overline{S_{v}}\ \overline{D^{ii}_{v}} = \sum_{j\neq i} \overline{S^{ij}_{v}}\quad\forall i$$

where the $N_{k_{1}}(i)$ is the index set of which $k_{1}$ nearest neighbors of $x^{v}_{i}$ have the same label with $x^{v}_{i}$.

From the construction of the graph of MFA, we can see that MFA can explore both label information and local manifold structure. Therefore, the learned new metric (projection) can ensure the within-class compactness and class separability simultaneously. Inspired the MFA, we can get a multi-view version of MFA algorithm with the optimization objective function:
\begin{equation}
\begin{split}
J = \underset {W^{v}}{max}\ \sum_{v=1}^{m}{Tr((W^{v})^{T}X^{v}B_{v}(X^{v})^{T}W^{v}})\\
\text{with}\quad Tr((W^{v})^{T}X^{v}L_{v}(X^{v})^{T}W^{v})=c_{v}\quad \forall v=1...m
\end{split}
\end{equation}

Although multiple types of representations of the shapes are extracted by different approaches, they are highly related since they are descriptions of the same shape. In order to guarantee the consensus principle, we make the hypotheses that the pairwise similarities of the representations are similar across all types of them. The hypotheses means that similarity matrixes of multiple types of representations should be consistent under the new metrics. Therefore, we employ the Hilbert-Schmidt Independence Criterion (HSIC) \cite{Gretton2005Measuring} as a measurement of dependence between the multiple types of shape representations.
In the MFAMML, the HSIC for any two types of representation can be simplified as:
\begin{equation}
\begin{split}
&HSIC(X^{v},X^{t}) =(N-1)^{-2}Tr(K_{v}HK_{t}H)\\
 & =  (n-1)^{-2}Tr((W^{v})^{T}X^{v}HK_{t}H(X^{v})^{T}W^{v})
 \label{Eq:7}
\end{split}
\end{equation}
where  the inner product kernel function is as:
$$K_{v} = (X^{v})^{T}W^{v}(W^{v})^{T}X^{v}.$$
In order to ensure our hypotheses, we use HSIC about any two types of representations to penalize for independence between them, which can make them to be consistent. Combine this with the MFA objectives for individual views Eq.\ref{Eq:7}, we can get the following the objective function:
\begin{equation}
\begin{split}
J = \underset {W^{v}}{max}\ \sum_{v=1}^{m}{Tr((W^{v})^{T}X^{v}B_{v}(X^{v})^{T}W^{v}}) \\
 +\lambda\sum_{w\neq i}^{m} Tr((W^{v})^{T}X^{v}HK_{w}H(X^{v})^{T}W^{v})\\
s.t. \quad Tr((W^{v})^{T}X^{v}L_{v}(X^{v})^{T}W^{v})=c_{v}
\quad \forall v=1...m
\label{Eq:8}
\end{split}
\end{equation}
where the $\lambda > 0$ is a a trade-off parameter between two terms of Eq[ref]. The first term ensure the class separability within each view for the most discrimination while the second one follows the hypotheses. We can find that seeking each projection $W^{v}$ exploits all other $W^{t},\ t\neq v$. Therefore, maximizing Eq\ref{Eq:8} aims to find the new projections for every view, which can integrate the geometric information from multiple types of representation.

\subsection{The Optimization Procedure Of MAFMML}
In this section, we introduce the details of optimizing the MFAMML. In order to find the optimal solution of objective function, we rewrite the objective function firstly:
\begin{equation}
\begin{split}
J  &=\ \underset {W^{v}}{max}\ \sum_{v=1}^{m}{Tr((W^{v})^{T}X^{v}B_{v}(X^{v})^{T}W^{v}}) \\
&+\lambda\sum_{w\neq i}^{m} Tr((W^{v})^{T}X^{v}HK_{w}H(X^{v})^{T}W^{v})\\
& = \underset {W^{v}}{max}\ \sum_{v=1}^{m}{Tr((W^{v})^{T}P_{v}W^{v})}\\
&s.t. \quad Tr((W^{v})^{T}X^{v}L_{v}(X^{v})^{T}W^{v})=c_{v}
\quad \forall v=1...m
\end{split}
\end{equation}
where
\begin{equation}
\begin{split}
&P_{v} = F + \lambda G_{v}, \quad
F = X^{v}B_{v}(X^{v})^{T},\quad \\
&G_{v} = \sum_{w=1;w\neq v}^{m} X^{v}HK_{w}H(X^{v})^{T}.
\label{Eq:10}
\end{split}
\end{equation}

The maximization problem is transforms into an iterative alternating optimization problem. First, we initialize all $W^{v},\ v=1,...,m$ by the solution of MFA for single view. We solve the objective of Eq\ref{Eq:10} by optimizing each of single $W^{t}$ at one time, keeping all other variables fixed. The optimization problem for  $W^{v}$ is as:
\begin{equation}
\begin{split}
 &\underset {W^{v}}{max}\ Tr((W^{v})^{T}P_{v}W^{v})\\
 &s.t. \quad Tr((W^{v})^{T}X^{v}L_{v}(X^{v})^{T}W^{v})=c_{v}. \\
 \label{Eq:11}
 \end{split}
\end{equation}
Obviously, optimization Eq.\ref{Eq:11} can be converted into the following equation:
\begin{equation}
\begin{split}
&\underset {W^{v}}{max}\ \frac{ Tr((W^{v})^{T}P_{v}W^{v})}
 {Tr((W^{v})^{T}X^{v}L_{v}(X^{v})^{T}W^{v})}
 \label{Eq:12}
\end{split}
\end{equation}
 Eq.\ref{Eq:12} is a standard Trace Ratio problem. Therefore, all the projection matrixes can be updated using Eq.\ref{Eq:12} respectively, which can be solved easily by Generalized Eigenvalue Decomposition (GED) algorithm. The iterative alternating optimization of MFAMML stops when all projection matrixes $W^{v}$ converges. Once $W^{v}$s are learned, we can obtain the new metrics $M_{v} = W^{v}(W^{v})^{T}$ for all types of the representations. Then we can compute the distance between every pair of the shapes as:
 \begin{equation}
 d(i,j) = \sum^{m}_{v=1} (x^{v}_{i}-x^{v}_{j})^{T}M_{v}(x^{v}_{i}-x^{v}_{j}).
 \end{equation}

\section{Experiments}
We conduct experiments on SHREC'10 datasets and compare out method with state-of-the-art methods. The standard of evaluation is based on these quantitative measure from PSB \cite{1314504}: Nearest Neighbor (NN), First Tier (FT), Second
Tier (ST), Emeasure (E), and Discounted Cumulative Gain (DCG).

\subsection{Parameters Setting}
For these two 3D shape datasets, we extract ShapeDNA, $HKS$, $WKS$, $siHKS$ spectral descriptors to form multiple types of shape representation. The ShapeDNA is used as the representation directly, other point descriptors are coded by BoW to form global representations.
 These spectral descriptors is shown as follows:
\begin{itemize}
	\item (1) ShapeDNA \cite{ReuterWP06}: The ShapeDNA is obtained by truncating the
	sequence of the eigenvalues of the LBO. Then it is normalized by the first eigenvalue. ShapeDNA is a global descriptor which has various properties such as the simple representation, comparison, scale invariance, ect. And in spite of its
	simplicity, it has a good performance for non-rigid shape retrieval.
	\item (2) $HKS$  \cite{SunOG09}: The $HKS$ is a point descriptor, which is defined as the diagonal element of the heat kernel based on the concept of heat diffusion over a surface, which represent that the amount of heat remaining at the point over a period of time. It has many advantages properties, such as intrinsic, multi-scale, informative and robust property.
	\item $siHKS$ \cite{BronsteinK10}: The $siHKS$ is a point descriptor, which is constructed based on $HKS$. The Fourier transform is utilized  for moving scale factor from the time domain to the frequency domain. Therefore, $siHKS$ is invariant for the change of the shape scale.
	\item $WKS$ \cite{AubrySC11}: The physical significance of $WKS$ is the average probability of a	quantum mechanical particle at a specific location. The $WKS$ clearly separates influences of different frequencies, treating all frequencies
	equally. Therefore, the $WKS$ can take more geometric details than $HKS$.
\end{itemize}

The first 35 normalized eigenvalues of the LBO are utilized as the ShapeDNA. Next, we explore first 100 eigenvectors and eigenvalues of LBO to compute the $HKS$, $siHKS$ and $WKS$ of every point for all the shapes. 50-dimensional $HKS$, $siHKS$ and 100-dimensional $WKS$ are obtained with same parameters setting in \cite{ChiotellisTWC16}. Then, we utilize the BoW algorithm to construct the three types of representations based on three point signatures, and these representations are all 64-dimensional.

\subsection{Experiment On SHREC'10}
\begin{figure}[!htb]
	\centering
	\includegraphics[width=1\linewidth]{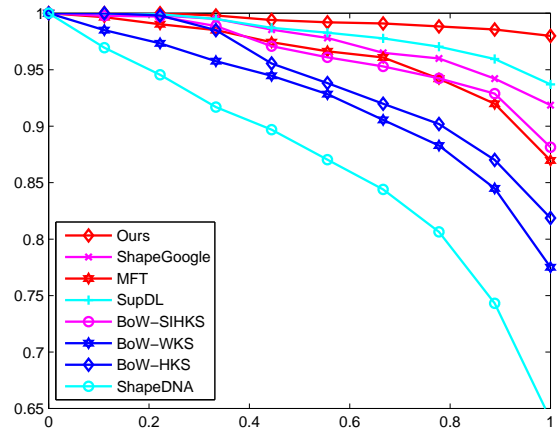}
	\caption{Comparison of the precision Recall curves (PR-curves) among our
		method and the other methods on SHREC10 Non-rigid dataset.}
	\label{fig:2}	
\end{figure}

SHREC'10 batabase \cite{Lian:2010:STN:2381147.2381167} consists of 200 watertight 3D triangular meshes. The dataset is classified into 10 categories based on their semantic (isometric), and each class contains 20 objects with distinct postures. To evaluate the performance of our algorithm, 2 experiments on SHREC'10 dataset are conducted. We compare our approach with the methods related with LBO: (1) ShapeGoogle, (2) Modal Function Transformation (MFT), (3)Supervised Dictionary Learning (SupDL), and these four representations without our algorithm. 50\% samples are randomly selected to train MFAMML and left 50\% samples are as testing set. The PR(precision-recall)-curves are shown in Fig.\ref{fig:2}. For next experiment, we compare our approach with these methods: MR-BF-DSIFT-E, DMEVD and CF, which are summarized in \cite{Lian:2010:STN:2381147.2381167}. In this experiment, we select 50\% samples randomly to train MFAMML also, and utilize all the dataset as the testing set for comparing. Table \ref{tab:1} shows the comparison between our methods and the methods summarized in \cite{Lian:2010:STN:2381147.2381167}.

\begin{table}[thp]\footnotesize
	\centering
	\caption{Five quantitative measures on SHREC'11} \label{tab:1}
	\addtolength{\tabcolsep}{1.5pt}
	\begin{tabular*}{7.95cm}{cccccc}
		\toprule[0.75pt]
		\hline
		Method &NN &FT &ST &E &DCG \\ \hline		
		MR-BF-DSIFT-E &98.5 &90.9 &96.3 &70.6.9 &97.6\\
		\hline
		DMEVD &100.0 &86.3 &95.7 &70.1 &97.7\\
		\hline
		CF &92.0 &63.5 &78.0 &55.3 &87.8\\
		\hline		
		\textbf {Our method} & \textbf{100.0} & \textbf{99.5} & \textbf{99.7} & \textbf{73.5}  & \textbf{99.8}\\ \hline	
		\bottomrule[0.75pt]
	\end{tabular*}
\label{Tab:1}
\end{table}

\subsection{Experiment Results}

 We can see that our performance out-performs the ShapeDNA, ShapeGoogle, SupDL, and the single representations from Fig.\ref{fig:2} . We then compare the methods summarized in SHREC'10. We see that our results is comparable to the best performing on these two datasets from Table\ref{tab:1}. Our methods obtain the best performance among the state-of-the-art methods on the common two datasets. These results verify the performance of the MFAMML method.

\section{Conclusion}
In this paper, we proposed a novel multi-view metric learning method for non-rigid 3D shape retrieval. MAFMML aims to overcome that one type of representation can not contains enough geometric information for retrieval task. The MAFMML can explore compatible and complementary geometric information from multiple intrinsic representations. For each type of representation, MFAMML learns a new metric for every representation by maximizing the distance between extra-class pair data points with inner-class pair fixed. Meanwhile, the HSIC is used to ensure the consistence across different types of representation. Therefore, MAFMML can integrates compatible information from multiple types of representations. Our approach is simple to implement and easy to optimize. Results of experiments on SHREC'10 benchmark datasets have proved the advantages of our approach over state-of-the-art non-rigid 3D shape retrieval methods.


\bibliographystyle{IEEEbib}
\bibliography{icme2019template}

\end{document}